\documentclass[letterpaper]{article} 
\usepackage[]{aaai23}  
\usepackage{times}  
\usepackage{helvet}  
\usepackage{courier}  
\usepackage[hyphens]{url}  
\usepackage{graphicx} 
\graphicspath{./}
\def\UrlFont{\rm}  
\usepackage{natbib}  
\usepackage{caption} 
\usepackage{algorithm}
\usepackage{algorithmic}

\usepackage{newfloat}
\usepackage{listings}
\DeclareCaptionStyle{ruled}{labelfont=normalfont,labelsep=colon,strut=off} 
\floatstyle{ruled}
\newfloat{listing}{tb}{lst}{}
\floatname{listing}{Listing}
\usepackage{amsfonts}
\usepackage{gensymb}

\title{Neural Implicit Surface Reconstruction from Noisy Camera Observations}
\author {
    Sarthak Gupta,\textsuperscript{\rm 1}
    Patrik Huber, \textsuperscript{\rm 2}
}
\affiliations {
    \textsuperscript{\rm 1} Indian Institute of Technology Roorkee, Roorkee, Uttarakhand, India - 247667\\
    \textsuperscript{\rm 2} University of York, Deramore Lane, Heslington, York, YO10 5GH, United Kingdom\\
    mrsarthakgupta@gmail.com, patrik.huber@york.ac.uk
}

\begin{document}

\maketitle

\begin{abstract}
Representing 3D objects and scenes with neural radiance fields has become very popular over the last years. Recently, surface-based representations have been proposed, that allow to reconstruct 3D objects from simple photographs.
However, most current techniques require an accurate camera calibration, i.e. camera parameters corresponding to each image, which is often a difficult task to do in real-life situations. To this end, we propose a method for learning 3D surfaces from noisy camera parameters. We show that we can learn camera parameters together with learning the surface representation, and demonstrate good quality 3D surface reconstruction even with noisy camera observations.

\end{abstract}

\section{Introduction}

The idea of representing 3D objects and scenes with neural networks, instead of traditional mesh-like representations, has gained significant traction recently. In \cite{DBLP:conf/eccv/MildenhallSTBRN20}, an approach called \emph{NeRF} is proposed, Neural Radiance Fields, where a neural network is used together with a volumetric representation to learn the representation of a scene from a collection of calibrated multi-view 2D images.
However, a volumetric representation is not the best representation in many cases; for example, many objects like faces are better represented using surfaces. 
This was tackled in a follow-up work by \cite{DBLP:conf/nips/WangLLTKW21}, called \emph{NeuS}, who proposed to use neural implicit surfaces together with volume rendering for multi-view reconstruction. 
This year, \cite{DBLP:journals/corr/abs-2102-07064} addressed another shortcoming of classical NeRF's, by extending the NeRF method to work on image data when camera calibration data is not present. This is achieved by jointly estimating the scene representation and optimising for the camera parameters, and the authors have shown promising results for frontal-facing scenes.


In this work, we propose to marry the benefits of each of these approaches: We propose a method to learn a neural implicit surface based representation of objects from noisy camera observations. We show that the classical NeuS method fails to learn an object completely if camera parameters are not precise, whereas our approach succeeds.
Furthermore, what is often taken as ground-truth camera parameters come themselves from a multi-view reconstruction software like COLMAP~\cite{schoenberger2016sfm}, and thus likely with estimation errors. By making camera parameters learnable, the approach gains the ability to correct for these errors, and to potentially achieve better quality than methods that take camera parameters as given truth.


\begin{figure}
    \centering
    \includegraphics[scale=0.25]{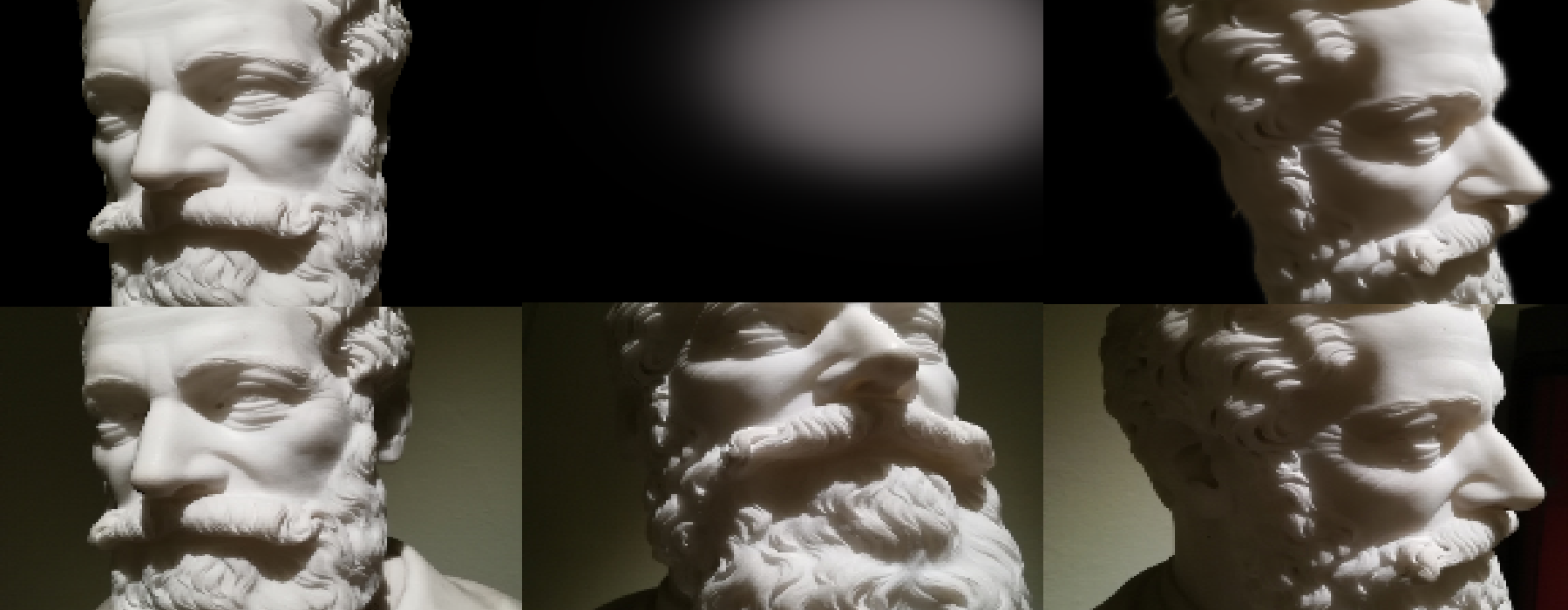}
    \includegraphics[scale=0.25]{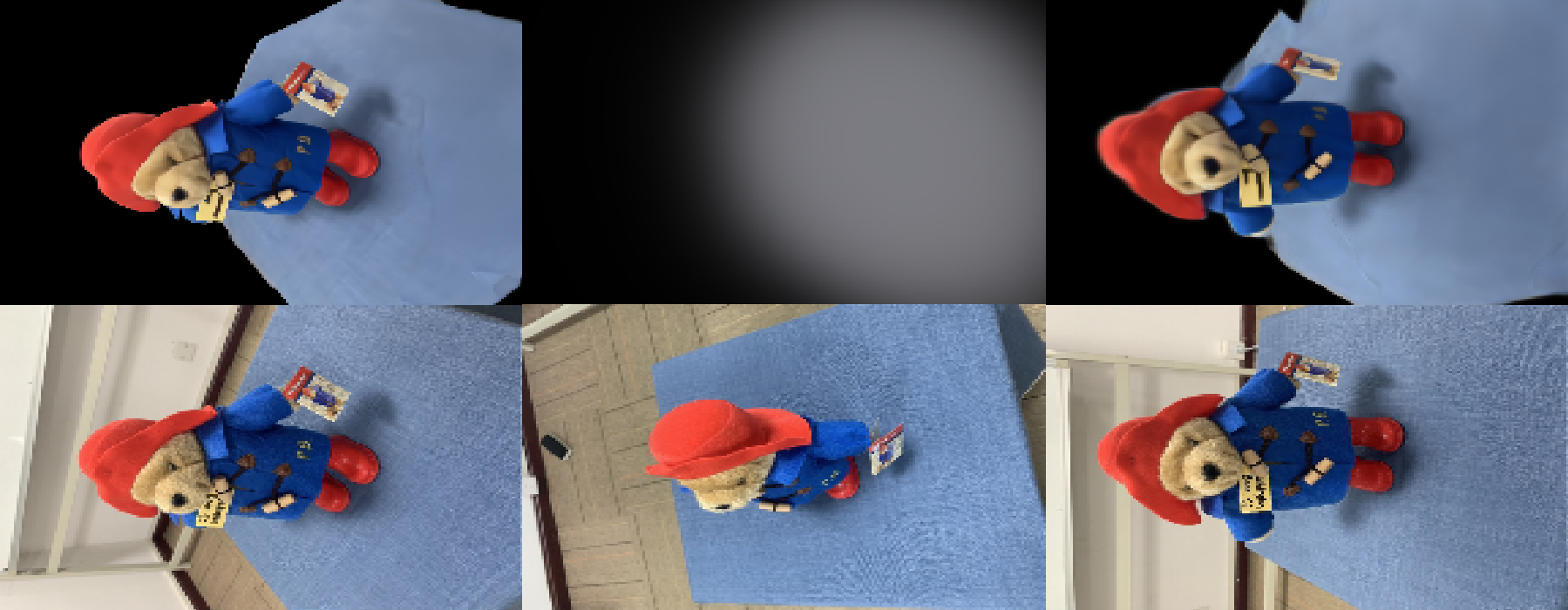}
    \caption{\emph{(Left)}: Reconstruction using \cite{DBLP:conf/nips/WangLLTKW21} with ground truth camera parameters. \emph{(Centre):} Reconstruction using \cite{DBLP:conf/nips/WangLLTKW21} with noisy camera parameters, where the approach completely fails. \emph{(Right):} Reconstruction using the proposed approach with noisy camera parameters. For each of the two objects, the lower images represent the actual image of the object while the upper image is the rendered image of the reconstructed surface.}
    \label{fig:image_results}
\end{figure}

\section{Methodology}

We draw inspiration from \cite{DBLP:journals/corr/abs-2102-07064}, where the authors make the camera parameters of a NeRF learnable to allow them to converge to values that result in desirable 3D scene reconstructions. While \cite{DBLP:journals/corr/abs-2102-07064} works with unknown camera parameters for only forward-facing input images with rotational and translational perturbations of up to  $\pm 20^{\circ}$, our approach works successfully on images from $360^{\circ}$ view angles. The latter results in a problem that is much harder to solve and is prone to local optima that do not produce good results when trying to learn the camera parameters from scratch.

The implicit surface based network of \cite{DBLP:conf/nips/WangLLTKW21}, \emph{NeuS}, consists of two MLPs to encode a signed distance function (SDF) and colour, respectively. Based on that, 
we add two modules that make both the extrinsic and intrinsic camera parameters learnable (see the Supplementary Material heading \emph{Network} for the full network architecture).
The extrinsic parameters are expressed as a $4\times4$ camera-to-world space transformation matrix with $T_{wc} = [R|t]$ where $R \in SO(3)$ and $t \in \mathbb{R}^3$ denote the camera rotation and translation, respectively. The camera intrinsics include the focal length and the principal point, using a pinhole camera model. We take the sensor's centre as the principal point, and we assume that the same camera takes all input images. Thus, estimating the camera intrinsics simplifies to finding the focal length.

The network, including the camera parameters, is trained on images of a scene using a weighted linear combination of Eikonal-loss \cite{DBLP:conf/icml/GroppYHAL20}, colour-loss, and mask-loss masking out the background. 
The loss is then back-propagated through the network, and back to the camera estimation modules. We further use separate optimisers for the learnable extrinsics and intrinsics, along with their own learning rate schedulers, to help the values to converge efficiently.




\section{Experiments \& Results}

\begin{table}[H]
    \centering
    \begin{tabular}{|c|c|c|}
    \hline
          \textbf{Case} & \textbf{Reconstruction loss} & \textbf{PSNR} \\
          \hline \hline
          bmvs\_man-\emph{baseline-gt} & 0.032 & 36.1 \\  
          \hline
          bmvs\_man-\emph{baseline-noisy} & 0.948 & 9.3 \\  
          \hline
          bmvs\_man-\emph{learnable-gt} & \textbf{0.028} & \textbf{37.5} \\
          \hline
          bmvs\_man-\emph{learnable-noisy} & 0.038 & 33.8 \\
          \hline \hline
          bmvs\_bear-\emph{baseline-gt} & 0.102 & 23.9 \\
          \hline
          bmvs\_bear-\emph{baseline-noisy} & 0.690 & 10.9 \\
          \hline
          bmvs\_bear-\emph{learnable-gt} & 0.115 & 23.4 \\
          \hline
          bmvs\_bear-\emph{learnable-noisy} & \textbf{0.098} & \textbf{26.2} \\
    \hline
    \end{tabular}
    \caption{We show results for two objects and four cases. Our approach with \emph{learnable} camera parameters performs as well or better than the \emph{baseline} (vanilla NeuS), whereas vanilla NeuS completely fails if camera parameters are noisy.}
    \label{tab:bmvs_loss_psnr_results}
\end{table}

We consider four cases to evaluate the performance of our method: We compare vanilla NeuS (\emph{baseline}) to our approach with \emph{learnable} camera parameters, both with ground-truth (\emph{gt}) and with noisy camera parameters (\emph{noisy}). We evaluate our approach quantitatively and qualitatively on two objects from the BlendedMVS dataset~\cite{DBLP:conf/cvpr/0008LLZRZFQ20}, \emph{bmvs\_bear} and \emph{bmvs\_man}.

For the extrinsics parameters, we sample noise from a normal distribution with a mean of zero and a standard deviation of $0.1$. While for the intrinsics parameters, we add noise with a mean of zero and standard deviation equal to $0.2$ times the mean value of the learnable parameters to make the noise of the same order as the values in the intrinsics matrix. Under these noise levels, we run the training process for 300,000 iterations while using the same hierarchical sampling as in \cite{DBLP:conf/nips/WangLLTKW21}.

We find that the baseline NeuS completely fails, as soon as camera parameters are not accurate. In the noisy camera parameter setting, our method produces reconstructions that are visually indistinguishable from that of baseline NeuS and produces a final reconstruction loss and Peak Signal-to-Noise Ratio (PSNR) similar to the baseline with an equal number of iterations.
Table~\ref{tab:bmvs_loss_psnr_results} summarises the results, and Figure~\ref{fig:image_results} shows example renderings for two objects.


\section{Conclusions \& Future Work}
While comparing our work to the current 3D surface reconstruction state-of-the-art in a noisy parameter setting, we have found that the existing method fails as soon as camera parameters are not accurate. Our method outperforms the existing method by being able to learn an equal or more accurate object representation, even in the presence of significant noise in the camera parameters.
Our work broadens the use cases of neural implicit surface based object reconstruction, by removing the need for accurate camera calibration information, and increasing the robustness to errors.

However, there is still much scope for work in this field, as reconstructing the surface from completely unknown camera parameters is still an open problem for $360^{\circ}$ view angles. To tackle this, we would like to investigate an approach akin to
\cite{DBLP:conf/iccv/JeongACACP21}, who learn a volumetric NeRF from unknown camera parameters. Furthermore, we plan to apply our method to a multi-view reconstruction benchmark, where 3D shape accuracy is evaluated.


\appendix

\bibliography{bibliography}

\begin{thebibliography}{7}
\providecommand{\natexlab}[1]{#1}

\bibitem[{Gropp et~al.(2020)Gropp, Yariv, Haim, Atzmon, and
  Lipman}]{DBLP:conf/icml/GroppYHAL20}
Gropp, A.; Yariv, L.; Haim, N.; Atzmon, M.; and Lipman, Y. 2020.
\newblock Implicit Geometric Regularization for Learning Shapes.
\newblock In \emph{{ICML}}, volume 119 of \emph{Proceedings of Machine Learning
  Research}, 3789--3799. {PMLR}.

\bibitem[{Jeong et~al.(2021)Jeong, Ahn, Choy, Anandkumar, Cho, and
  Park}]{DBLP:conf/iccv/JeongACACP21}
Jeong, Y.; Ahn, S.; Choy, C.~B.; Anandkumar, A.; Cho, M.; and Park, J. 2021.
\newblock Self-Calibrating Neural Radiance Fields.
\newblock In \emph{{ICCV}}, 5826--5834. {IEEE}.

\bibitem[{Mildenhall et~al.(2020)Mildenhall, Srinivasan, Tancik, Barron,
  Ramamoorthi, and Ng}]{DBLP:conf/eccv/MildenhallSTBRN20}
Mildenhall, B.; Srinivasan, P.~P.; Tancik, M.; Barron, J.~T.; Ramamoorthi, R.;
  and Ng, R. 2020.
\newblock NeRF: Representing Scenes as Neural Radiance Fields for View
  Synthesis.
\newblock In \emph{{ECCV} {(1)}}, volume 12346 of \emph{Lecture Notes in
  Computer Science}, 405--421. Springer.

\bibitem[{Sch\"{o}nberger and Frahm(2016)}]{schoenberger2016sfm}
Sch\"{o}nberger, J.~L.; and Frahm, J.-M. 2016.
\newblock Structure-from-Motion Revisited.
\newblock In \emph{Conference on Computer Vision and Pattern Recognition
  (CVPR)}.

\bibitem[{Wang et~al.(2021{\natexlab{a}})Wang, Liu, Liu, Theobalt, Komura, and
  Wang}]{DBLP:conf/nips/WangLLTKW21}
Wang, P.; Liu, L.; Liu, Y.; Theobalt, C.; Komura, T.; and Wang, W.
  2021{\natexlab{a}}.
\newblock NeuS: Learning Neural Implicit Surfaces by Volume Rendering for
  Multi-view Reconstruction.
\newblock In \emph{NeurIPS}, 27171--27183.

\bibitem[{Wang et~al.(2021{\natexlab{b}})Wang, Wu, Xie, Chen, and
  Prisacariu}]{DBLP:journals/corr/abs-2102-07064}
Wang, Z.; Wu, S.; Xie, W.; Chen, M.; and Prisacariu, V.~A. 2021{\natexlab{b}}.
\newblock NeRF{-}{-}: Neural Radiance Fields Without Known Camera Parameters.
\newblock \emph{CoRR}, abs/2102.07064.

\bibitem[{Yao et~al.(2020)Yao, Luo, Li, Zhang, Ren, Zhou, Fang, and
  Quan}]{DBLP:conf/cvpr/0008LLZRZFQ20}
Yao, Y.; Luo, Z.; Li, S.; Zhang, J.; Ren, Y.; Zhou, L.; Fang, T.; and Quan, L.
  2020.
\newblock BlendedMVS: {A} Large-Scale Dataset for Generalized Multi-View Stereo
  Networks.
\newblock In \emph{{CVPR}}, 1787--1796. Computer Vision Foundation / {IEEE}.

\end{thebibliography}


\end{document}


\maketitle

\section{Dataset}

The images used in our experiments are from the BlendedMVS (\emph{bmvs}) dataset~\cite{DBLP:conf/cvpr/0008LLZRZFQ20}(CC-4 License).
The two tested scenes, \emph{bmvs\_man} and \emph{bmvs\_bear}, are scenes from the challenging low-res set of the BlendedMVS dataset. Each scene consists of 31 to 143 images with a resolution of 768 × 576 pixels, that are arranged in a 360\degree circle around the object. The per-pixel masks provided by the BlendedMVS dataset are used to mask the background during the NeuS training.

\begin{figure}[h]
    \centering
    \includegraphics[width=1.0\columnwidth]{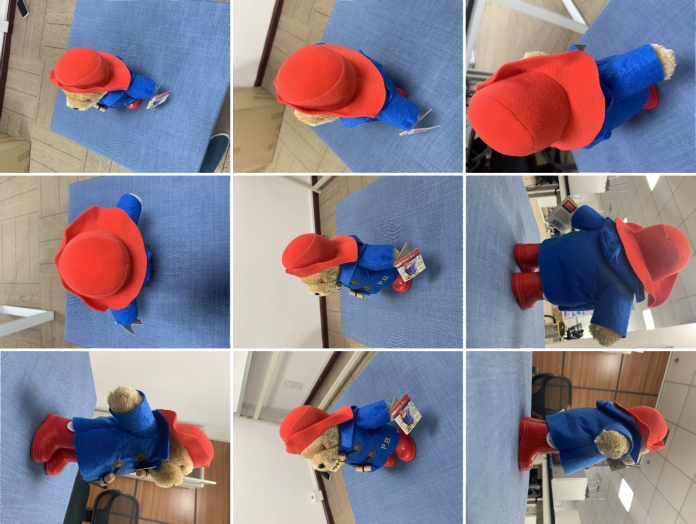}
    \caption{A subset of the 123 images used to train bmvs\_bear.}
    \label{fig:bmvs_bear_data_subset_small}
\end{figure}

Figure~\ref{fig:bmvs_bear_data_subset_small} shows a subset of the 123 images used to train bmvs\_bear.

\section{Network}

For our \emph{baseline} experiments, we use the original architecture and code from \cite{DBLP:conf/nips/WangLLTKW21}. The network consists of two MLPs to encode the signed distance function (SDF) and colour respectively. The signed distance function is modelled with an MLP that consists of 8 hidden layers with a hidden size of 256, and Softplus activation with $ \beta = 100 $ for all hidden layers. A skip connection additionally connects the input with the output of the fourth layer. The colour prediction network is modelled with an MLP with 4 hidden layers of size 256. It takes as inputs the spatial location, the view direction, normal vector of the SDF, and a 256-dimensional feature vector from the SDF MLP. A positional encoding is applied to both the spatial location input and the view direction.

\section{Training}

\begin{figure}[h]
    \centering
    \includegraphics[scale=0.65]{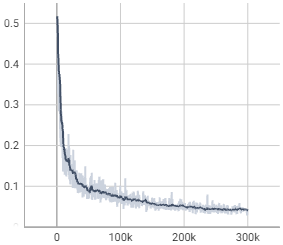}
    \caption{Loss plot from the training of the experiment bmvs\_man-\emph{learnable-noisy}.}
    \label{fig:bmvs_man__learnable_noisy__loss_plot}
\end{figure}

All networks are trained for 300,000 epochs, on an NVIDIA Quadro P6000 GPU. Figure~\ref{fig:bmvs_man__learnable_noisy__loss_plot} shows the loss plot from the training of the experiment bmvs\_man-\emph{learnable-noisy}, showing stable learning and convergence.

\section{Additional Results}

\begin{figure}[htbp]
    \centering
    \includegraphics[width=1.0\columnwidth]{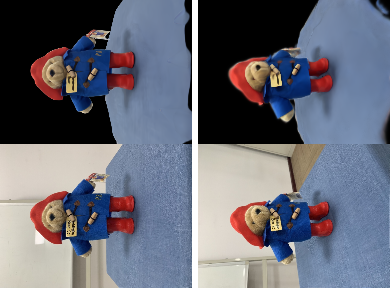}
    \caption{\emph{(left):} Result of vanilla NeuS, trained with ground-truth camera parameters (\emph{baseline-gt}). \emph{(right):} Our result with learnable camera parameters, learned from very noisy initialisation (\emph{learnable-noisy}). The top half shows the neural network rendering, the bottom half shows the respective ground truth camera photo.}
    \label{fig:bmvs_bear_closeup}
\end{figure}

\begin{figure}[htbp]
    \centering
    \includegraphics[width=1.0\columnwidth]{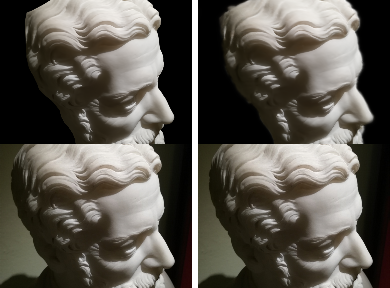}
    \caption{\emph{(left):} Result of vanilla NeuS, trained with ground-truth camera parameters (\emph{baseline-gt}). \emph{(right):} Our result with learnable camera parameters, learned from very noisy initialisation (\emph{learnable-noisy}). The top half shows the neural network rendering, the bottom half shows the respective ground truth camera photo.}
    \label{fig:bmvs_man_closeup}
\end{figure}

In Figures~\ref{fig:bmvs_bear_closeup} and~\ref{fig:bmvs_man_closeup}, we show close-ups of bmvs\_bear and bmvs\_man respectively. On the left side, we show the result of vanilla NeuS, trained with ground-truth camera parameters (\emph{baseline-gt}). On the right side, we show our result with learnable camera parameters, learned from very noisy initialisation (\emph{learnable-noisy}). On all images, the top half shows the neural network rendering, whereas the bottom half shows the respective ground truth camera photo.

We don't depict the results for \emph{baseline-noisy}, as the method in these cases was not able to learn the scene successfully.

\appendix

\bibliography{bibliography}
